\documentclass{article}

\usepackage[final]{neurips_2025}

\usepackage[utf8]{inputenc}
\usepackage[T1]{fontenc}
\usepackage{hyperref}
\usepackage{url}
\usepackage{booktabs}
\usepackage{diagbox}
\usepackage{amsfonts}
\usepackage{nicefrac}
\usepackage{microtype}
\usepackage{xcolor}
\usepackage{amsmath,amsthm}
\usepackage{adjustbox}
\usepackage{enumitem}
\usepackage{multirow}
\usepackage{rotating}

\usepackage{hyperref}
\usepackage{url}

\usepackage[utf8]{inputenc}
\usepackage[T1]{fontenc}
\usepackage{hyperref}
\usepackage{url}
\usepackage{booktabs}
\usepackage{amsfonts}
\usepackage{nicefrac}
\usepackage{microtype}
\usepackage{xcolor}

\usepackage{comment}
\usepackage{amsmath}
\usepackage{graphicx}
\usepackage{subcaption}
\usepackage[ruled,vlined,linesnumbered,noend]{algorithm2e}
\usepackage{amsmath}
\usepackage{amssymb}
\usepackage{mathtools}
\usepackage{amsthm}
\usepackage{pgfplots}
\usepackage{microtype}
\usepackage{graphicx}
\usepackage{booktabs}
\pgfplotsset{compat=1.18} 
\usepackage{multirow}

\usepackage{amssymb}
\usepackage{pifont}

\usepackage{enumitem}
\usepackage{accents}
\usepackage{caption}
\usepackage{graphicx}

\usepackage[mathscr]{euscript}
\usepackage{cleveref}
\usepackage{wrapfig}

\usepackage{nicematrix}

\newcommand{\augU}{{\widehat{U}}}
\newcommand{\augS}{{\widehat{S}}}

\newcommand{\augV}{{\widehat{V}}}

\newcommand{\norm}[1]{\left\lVert#1\right\rVert}

\newcommand{\rom}[1]{\uppercase\expandafter{\romannumeral #1\relax}}

\newcommand{\regLoss}{\widetilde{\mathcal{L}}}
\newcommand{\loss}{{\mathcal{L}}}

\def\ALG{DLRT}

\title{Compressing Vision Transformers in Geospatial Transfer Learning with Manifold-Constrained Optimization}

\author{%
 Thomas Snyder \\
 Yale University \\
  201 York Street, New Haven, CT 06511, USA \\
  \texttt{thomas.snyder@yale.edu} \\
   \And
   H. Lexie Yang \\
   Geospatial Science and Human Security Division \\
   Oak Ridge National Laboratory \\
   Oak Ridge, TN 37831 USA \\
   \texttt{yangh@ornl.gov} 
   \And
   Stefan Schnake \\
   Computer Science and Mathematics Division \\
   Oak Ridge National Laboratory \\
   Oak Ridge, TN 37831 USA \\
   \texttt{schnakesr@ornl.gov}
   \And
     Steffen Schotth\"ofer \\
   Computer Science and Mathematics Division \\
  Oak Ridge National Laboratory \\
  Oak Ridge, TN 37831 USA \\
   \texttt{schotthofers@ornl.gov}
}

\begin{document}

\maketitle

\begin{abstract}
Deploying geospatial foundation models on resource-constrained edge devices demands compact architectures that maintain high downstream performance. However, their large parameter counts and the accuracy loss often induced by compression limit practical adoption.
In this work, we leverage manifold-constrained optimization framework DLRT to compress large vision transformer–based geospatial foundation models during transfer learning. By enforcing structured low-dimensional parameterizations aligned with downstream objectives, this approach achieves strong compression while preserving task-specific accuracy. We show that the method outperforms of-the-shelf low-rank methods as LoRA. Experiments on diverse geospatial benchmarks confirm substantial parameter reduction with minimal accuracy loss, enabling high-performing, on-device geospatial models.
\end{abstract}

\section{Introduction}

The proliferation of high-resolution remote-sensing imagery and advances in large-scale training have driven rapid progress in geospatial representation learning for applications as mapping of built environments \cite{9337930}, disaster management \cite{osti_1671424}, and gravity anomaly mapping \cite{yang_desasters}. Recent work \cite{dias2024oreole} shows clear benefits from building large foundation models (FMs) and scaling vision-transformer (ViT) architectures to hundreds of millions or billions of parameters for remote sensing tasks, yielding strong generalization across multiple downstream tasks, including classification, segmentation and object detection. In practice, geospatial applications often begin with large FMs, which can then be fine-tuned through transfer learning to support specific use cases such as land-cover mapping, disaster response, and near-real-time monitoring. In operational and deployment environment, practitioners require models that (i) meet limited memory and latency constraints, and (ii) preserve task-specific accuracy and calibration. While these models enable strong transfer performance, their large size and computational demands high operational costs, making serving such large FM based models impractical for many practitioners for edge or local deployment, where limited compute and memory resources are available \cite{schotthöfer2024federateddynamicallowranktraining}. Model compression methods often invoke an often ambiguous trade-off between parameter reduction and reliable downstream performance.

Recently, principled methods to impose a structured low-rank parameterization during training for model compression have been proposed. Dynamical low-rank training (DLRT) approaches \cite{schotthoefer2022lowrank, schotthöfer2024federateddynamicallowranktraining, schotthoefer2025momentum} restrict weight updates to a low-dimensional manifold; this yields both practical memory savings and provable approximation/descent guarantees under standard assumptions. 

In this paper we present a careful, large-scale empirical study that applies manifold-constrained DLRT methods to compress ViT-based geospatial foundation models \emph{during} supervised transfer learning. Our study targets both ImageNet21k–initialized ViTs \cite{dosovitskiy2021imageworth16x16words} and the medium-resolution (MR) version of OReole geospatial FMs \cite{Dias_2024}, and focuses on land-cover classification tasks on standard benchmarks.

\section{Recap: Dynamical Low-Rank Training}
We consider a neural network $f$ as a concatenation of $L$ layers
$
z^{\ell+1}=\sigma^\ell(W^{\ell}z^\ell)
$
with matrix valued\footnote{Extensions to tensor-valued layers, e.g. in CNNs, are available \cite{NEURIPS2024_ea48cb23}} parameters $W^\ell\in\mathbb{R}^{n\times n}$, layer input $z^\ell\in\mathbb{R}^{n\times b}$ and element-wise nonlinear activation $\sigma^\ell$. 
The data $X$ constitutes the input to the first layer, i.e. $z^0=X$. 
The network is trained on a loss function $\loss$ which we assume to be locally bounded with a Lipschitz continuous gradient. Throughout this work, we call a network in the standard format a "baseline" network.\looseness=-1

\begin{wrapfigure}{R}{0.3\textwidth}
\vspace{-1.8em}
    \centering
    \includegraphics[width=\linewidth]{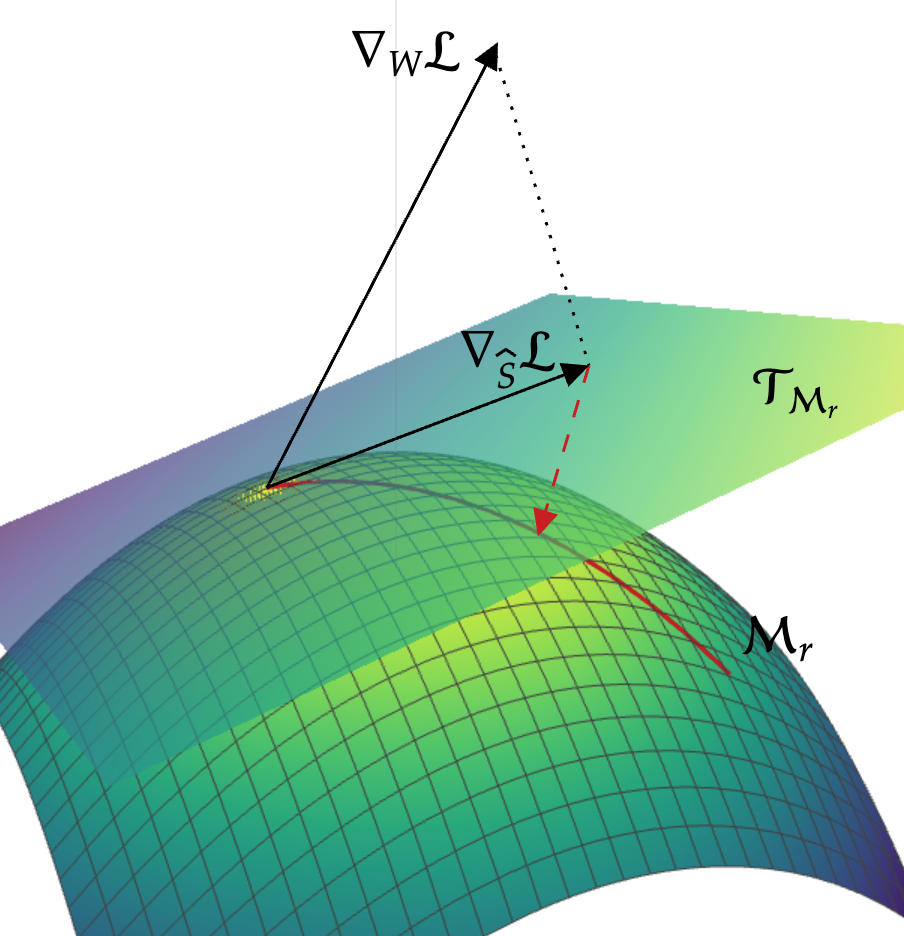}
      \caption{Geometric interpretation of DLRT. First, we compute the parametrization of the tangent plane $\mathcal{T}_{\mathcal{M}_r}$. Then we compute the projected gradient update with $\nabla_{\widehat{S}}\mathcal{L}$. Lastly, we retract the updated coefficients back onto the manifold $\mathcal{M}_r$. The regularizer $\mathcal{R}$ effectively changes the local curvature of $\mathcal{M}_r$.}
    \label{fig_manifold}
    \vspace{-1em}
\end{wrapfigure}
\textbf{Low-rank Compression:} {The compression of the network for training and inference} is typically facilitated by approximating the layer weight matrices by a low-rank factorization $W^\ell= U^\ell S^\ell V^{\ell,\top}$ with $ U^\ell,V^{\ell}\in\mathbb{R}^{n\times r}$ and $S^\ell\in\mathbb{R}^{r\times r}$, where $r$ is the rank of the factorization. For simplicity of exposition we drop the layer index $\ell$ and consider a single layer in the following. We denote the gradient flow of the layer by $\dot W(t)=-\nabla_W\loss(W(t))$.

\textbf{Manifold constrained low-rank training} If $\text{rank}(S)=r$, the factorization $USV^\top$ is element of the manifold of rank $r$ matrices $\mathcal{M}_r$. 
Recent work \cite{schotthoefer2022lowrank} interprets the low-rank training process as the evolution the gradient flow   projected onto $\mathcal{M}_r$, i.e., $\dot W_r(t)=-P_{\mathcal{M}_r}\nabla_{W_r}\loss(W(t))$. This point of view is useful to build efficient optimizers for low-rank layers for model compression with improved convergence behavior \cite{NEURIPS2024_ea48cb23,kusch2025augmentedbackwardcorrectedprojectorsplitting,Hnatiuk}, resource constrained federated learning \cite{schotthöfer2024federateddynamicallowranktraining}, low-rank finetuning \cite{schotthoefer2024geolora} and it has been extended to momentum-based optimizers \cite{schotthfer2025geometric}. 

Below we summarize the method, called \textit{dynamical low-rank training} (DLRT). 

\paragraph{Basis Augmentation:}
The method first augments the current bases $U^t,V^t$ at optimization step $t$ by their gradient dynamics $\nabla_U\mathcal{L}$, $\nabla_V\mathcal{L}$ via
\begin{align*}\label{eq_basis_augmentation}
\begin{aligned}
\augU = \texttt{orth}([U^t \mid \nabla_U\mathcal{L}(U^tS^tV^{t,\top})]),\qquad
\augV = \texttt{orth}([V^t \mid \nabla_V\mathcal{L}(U^tS^tV^{t,\top})])\in\mathbb{R}^{n\times 2r},
\end{aligned}
\end{align*}
to double the rank of the low-rank representation and subsequently creates orthonormal bases $\augU, \augV$.  
The span of $\augU$ contains $U^t$, which is needed to ensure that the loss does not increase during augmentation, and a first-order approximation of $\mathrm{span}(U^{t+1})$ using the exact gradient flow for $U$, see \cite[Theorem 2]{schotthöfer2024federateddynamicallowranktraining} for details.
Geometrically, the latent space 
\begin{equation}
\mathcal{S}=\{\augU Z\augV^\top:Z\in\mathbb{R}^{2r\times 2r}\}
\end{equation}
can be seen as subspace

of the tangent plane of $\mathcal{M}_r$ at $U^tS^tV^{t,\top}$, see \Cref{fig_manifold}. 

\paragraph{Latent Space Training:}
We update the latent coefficients $\augS$ via a Galerkin projection of the training dynamics onto the latent space $\mathcal{S}$.
The latent coefficients $\augS$ are updated by integrating the projected gradient flow  $\dot{\augS} = -\augU^{\top}\nabla_W\regLoss\augV =-\nabla_\augS\regLoss$ using stochastic gradient descent or an other suitable optimizer for a number of $s_*$ local iterations, i.e. 
\begin{align}\label{eq_coeff_update}
     {\augS_{s+1}} &=\augS_{s}-\lambda\nabla_\augS\mathcal{L}\quad s=0,\ldots,s_*-1. 
\end{align}

\Cref{eq_coeff_update} is initialized with ${\augS_{0}} = \augU^\top U^t S^t V^{t,\top} \augV\in\mathbb{R}^{2r\times 2r}$, and we set $\tilde{S}=\hat{S}_{s_*}$

\paragraph{Truncation:}
Finally, the latent solution $\augU\tilde{S}\augV^\top$ is retracted back onto the manifold $\mathcal{M}_r$. The retraction can be computed efficiently by using a truncated SVD of $\tilde{S}$ that discards the smallest $r$ singular values. 
To enable rank adaptivity, the new rank $r_1$ instead of $r$ can be chosen by a variety of criteria, e.g., a singular value threshold $\norm{[\varsigma_{r_1},\dots,\varsigma_{2r}]}_2<\vartheta$.
Once a suitable rank is determined, the bases $U$ and $V$ are updated by discarding the basis vectors corresponding to the truncated singular values.\looseness=-1  

\paragraph{Computational cost:} The computational cost of \ALG~is asymptotically the same as LoRA \cite{hu2021lora}, since the reconstruction of the full weight matrix $W$ is never required. The orthonormalization accounts for $\mathcal{O}(nr^2)$, the regularizer $\mathcal{R}$ for $\mathcal{O}(r^3)$, and the SVD for $\mathcal{O}(r^3)$ floating point operations.  When using multiple coefficient update steps $s_*>1$, the amortized cost is indeed lower than that of LoRA, since only the gradient with respect to $\augS$ is required in most updates.

\paragraph{Convergence:} Careful construction of the tangent space during the augmentation step guarantees loss descent of the DLRT scheme\cite{schotthoefer2024geolora}, i.e.,  
{\small
\begin{align*}
\mathbb{E}_{\xi_{t+1}}\!\left[ L\left(W_{r}^{t+1}\right) \right]
\le
L\left(W_{r}^{t}\right)
- \lambda \left( 1 - \frac{L \lambda^{2}}{2} \right)
\mathbb{E}_{\xi_{1}}\!\left[ \left\| P_{\mathcal{M}_r}\nabla_{W_r}\loss(W_{r}^{t},\xi_t) \right\|^{2} \right]
+ L\,\mathbb{E}_{\xi_{1}}\!\left[ \left\| W_{r}^{t+1} - \augU\tilde{S}\augV^\top \right\| \right],
\end{align*}
} where $L$ is the Lipschitz constant and $\xi_t$ denotes the stochastic influence of batch descent.
This is effectively the projected loss descent of stochastic gradient descent, except for the retraction error term 
$L\,\mathbb{E}_{\xi_{1}}\![ \| W_{r}^{t+1} - \augU\tilde{S}\augV^{\top} \| ]$. In practice, this term vanishes as a suitable manifold rank and basis is found \cite{Coquelin_2024}, paving the way for convergence of the scheme \cite[Theorem 2]{schotthoefer2024geolora} under standard assumptions, i.e., 
\begin{align*}
\liminf_{T \to \infty} \, \mathbb{E} \!\left[ \left\| P_{\mathcal{M}_r}\nabla_{W_r}\loss(W_{r}^{t},\xi_t)  \right\|^{2} \right] = 0
\end{align*}

\paragraph{Error control and robustness:} Orthogonality of the bases and the manifold constraint optimization enables error control of the low-rank scheme in relation to full-rank training. That is, the solution of the original gradient flow $W(t)$ and the projected (Riemannian) gradient flow $W_r(t)$ are close to each other, if they start from the same initial condition \cite[Theorem 3]{schotthoefer2024geolora}, 
\begin{align*}
\left\| W(t) - W^{r}_{t} \right\| \le c_{1}\varepsilon + c_{2}\lambda + \frac{c_{3}\vartheta}{\lambda},
\end{align*}
where $c_{1,2,3}>0$ do not depend on time or the geometry of the manifold. 

Lastly, the method is compatible with regularization schemes \cite{schotthöfer2025dynamicallowrankcompressionneural} acting on the low-rank latent space of the manifold representation to improve adversarial robustness, an important property for geospatial applications.

\section{Low-Rank Compressed Transfer Learning for Geospatial Vision Transformers}

\textbf{Test case setup}
We explore the capabilities of low-rank manifold training and concurrent automatic compression of geospatial foundation models. We target the image classification problem as the downstream task using three well-known remote sensing benchmark datasets, NWPU\cite{7891544}, AID\cite{Xia_2017}, and UCM\cite{10.1145/1869790.1869829}.  
The test cases have small train-to-validation set ratios of 50\% for UCM, 20\% for AID,  and 10\% for NWPU, rendering them challenging benchmarks for transfer learning. Test case and data preprocessing details are provided in \Cref{sec_test_case_descr}.

\textbf{Foundation models.} We consider both "classic" ViT models \cite{dosovitskiy2021imageworth16x16words} and the OReole-MR family of geospatial foundation models (FMs) \cite{Dias_2024}. The classic ViTs include ViT-B16 and ViT-H14, using HuggingFace checkpoints \texttt{google/vit-base-patch32-224-in21k} and \texttt{google/vit-huge-patch14-224-in21k}, pretrained on ImageNet-21k \cite{ridnik2021imagenet21kpretrainingmasses}. Architectural details are provided in \Cref{tab_vits}.

The OReole-MR models follow the ViT architecture but adopt a self-supervised pretraining scheme, embedding the ViT backbone within a masked autoencoder (MAE) pipeline \cite{dias2024oreole}. Pretraining is conducted on the unlabeled Million-AID dataset \cite{long2021creatingbenchmarkdatasetaerial}. For downstream tasks in this work, we extract the pretrained ViT backbone from the MAE and attach a task-specific classification head.

For a pretrained foundation model, we apply a truncated SVD of rank $r$ to each fully connected layer, obtaining $U,S,V$ factors that initialize low-rank training consistent with the pretrained weights. The classification head is excluded, as its rank equals the number of classes.

\begin{table}[t]
\caption{Validation accuracy comparison (in percent) of foundation models pretrained on ImageNet-21k with supervised pretraining, and Oreole style \cite{dias2024oreole} MAE self supervised pretraining. The models are compressed during transfer learning on the UCM, AID, and NWPU  downstream datasets. "c.r." denotes compression ratio, i.e. the percentage of removed parameters compared to baseline. Each cell denotes the mean and standard deviation over 10 runs with stochastic gradient descent. }
\label{tab:neurips_table}
\resizebox{\textwidth}{!}
{
\centering
\setlength{\tabcolsep}{4pt}
\renewcommand{\arraystretch}{1.2}
\begin{tabular}{ll
                *{5}{c}
                *{5}{c}
                *{5}{c}
                }
\toprule
&\multirow{2}{*}{\textbf{Model}} & 
\multicolumn{5}{c}{\textbf{UCM}} & 
\multicolumn{5}{c}{\textbf{AID}} & 
\multicolumn{5}{c}{\textbf{NWPU}} \\
\cmidrule(lr){3-7} \cmidrule(lr){8-12} \cmidrule(lr){13-17}
& & Baseline & DLRT & c.r. & LoRA & c.r. 
& Baseline & DLRT & c.r. & LoRA & c.r. 
& Baseline & DLRT & c.r. & LoRA & c.r. \\
\midrule
\multirow{3}{*}{\rotatebox[origin=c]{90}{{\shortstack{\textcolor{white}{ooo}Image\\\textcolor{white}{ooo}Net21k}}}} 
& ViT-B16 & 98.13$\pm$0.15 & 97.67$\pm$0.13 &68.63 &  95.37$\pm$0.34 & 62.38
        & 95.42$\pm$0.68 & 93.92$\pm$0.55& 66.15 & 93.35$\pm$0.45 & 66.13
        & 91.45$\pm$0.43 & 89.41$\pm$0.21 & 65.35 & 88.01$\pm$1.00 & 62.40 \\
& ViT-H14 & 97.57$\pm$0.89 & 97.32$\pm$0.42 & 81.47 &96.86$\pm$0.63 & 78.33
        & 93.75$\pm$0.39 &93.83$\pm$0.18 & 78.22 & 90.67$\pm$0.08 & 78.01
        & 90.04$\pm$0.73 & 89.69$\pm$0.57 &79.52 & 88.51$\pm$0.63 & 79.53 \\
\midrule
\multirow{3}{*}{\rotatebox[origin=c]{90}{\shortstack{OReole\\MR}}} 
& ViT-B16& 97.87$\pm$0.94 &96.74$\pm$0.18 & 64.32 &95.05$\pm$0.89 & 62.42
        & 95.01$\pm$0.43 & 90.12$\pm$0.83 & 70.13 &  88.02$\pm$0.98 & 62.35 
        & 91.35$\pm$1.10 & 84.22$\pm$0.20 & 63.78 & 81.06$\pm$0.77 & 63.41 \\
& ViT-H14 & 98.56$\pm$0.63 &97.17$\pm$0.67 & 80.83 & 94.28$\pm$0.78& 77.69 
        & 96.04$\pm$0.28 & 92.91$\pm$0.56 & 78.22 & 90.75$\pm$0.75 & 78.03
        & 92.79$\pm$0.24 & 88.80$\pm$0.80 & 78.22 & 88.01$\pm$0.21& 76.65 \\
&  ViT-G14 & 98.30$\pm$0.75 & 98.29$\pm$0.17 &77.33 & 94.00$\pm$0.91 & 81.96
        & 95.76$\pm$0.23 &93.85$\pm$0.67& 81.69 &  89.48$\pm$0.64 & 76.35
        & 92.27$\pm$0.20&  88.25$\pm$0.35 & 80.56 & 82.24$\pm$4.23 &79.65 \\
\bottomrule
\end{tabular}

}
\end{table}

\textbf{Transfer learning}:
We compare the compressed transfer learning capabilities of DLRT against two baselines: uncompressed transfer learning with full models, and training only the low-rank factors, akin to the popular LoRA method \cite{hu2021lora}. For each combination of training method, downstream dataset, and model, hyperparameters were chosen via an initial sweep (see \Cref{tab:combined_hyperparameters}). We find that the learning rate most strongly affects validation accuracy, with smaller values performing best, which consistent with transfer learning. In contrast, the DLRT hyperparameters $\tau$ and $s_*$ show an order of magnitude weaker correlation, indicating robustness to their choice.

For both LoRA and DLRT applied to the OReole-MR models, we found it beneficial to include a single warm-up epoch of standard (uncompressed) transfer learning before applying low-rank adaptation. This simple step improved final validation accuracy by up to $10\%$ compared to training without warm-up, a trend observed consistently across both DLRT and LoRA-based training. Interestingly, this effect was absent for ImageNet-pretrained checkpoints, which we attribute to the difference in pretraining schemes: the self-supervised pretraining of OReole-MR appears less aligned with the low-rank subspace, whereas the supervised pretraining of ImageNet models yields weights more naturally compatible with DLRT- or LoRA-style compression. A more detailed investigation of this phenomenon is left to future work.

\textbf{Results:}
Table \ref{tab:neurips_table} compares validation accuracy on UCM, AID, and NWPU for ViT models pretrained on ImageNet-21k with either supervised or Oreole-MR MAE pretraining. We evaluate baseline finetuning against DLRT and LoRA under comparable compression ratios.  Results are reported as the mean and standard deviation over $10$ runs.

DLRT consistently preserves accuracy within 0.5–1\% of the baseline in the UCM dataset, while removing 64–82\% of parameters, whereas LoRA exhibits larger drops in validation accuracy, especially for the OReole-MR models. 
In the more challenging AID and NWPU test cases with smaller train-to-test data ratio, the accuracy of low-rank method drops more compared to the baseline, however DLRT surpasses LoRA based training by 1-6\% validation accuracy. 
Across pretraining types and model scales, DLRT outperforms LoRA in accuracy retention under equal compression. 
Finally, we observe that OReole-MR based FMs achieve the highest validation accuracies across test cases, but are harder to compress in low-rank format, compared to the ImageNet-based models.

\section{Conclusion}

This study demonstrates that manifold-constrained low-rank optimization can substantially compress vision transformer–based geospatial foundation models while preserving performance of downstream tasks. By enforcing structured parameterizations aligned with transfer learning objectives, the approach consistently outperforms standard low-rank baselines such as LoRA across multiple datasets and pretraining schemes. These findings highlight the practicality of deploying compact, high-performing geospatial models on constrained hardware, paving the way for broader real-world applications. Future work may improve alignment with self-supervised pretraining, and extensions to additional geospatial tasks.

\bibliographystyle{abbrv}
\bibliography{ref}

\newpage
\appendix

\section{Test case descriptions}\label{sec_test_case_descr}

\subsubsection{UCM}

The University of California, Merced (UCM) Land Use Dataset is a benchmark dataset in remote sensing and computer vision, introduced in \cite{10.1145/1869790.1869829}. It covers urban areas across the United States. It comprises 2,100 high-resolution aerial RGB images, each measuring 256×256 pixels, categorized into 21 land use classes with 100 images per class. The images were manually extracted from the USGS National Map Urban Area Imagery collection, covering various urban areas across the United States. The dataset contains images with spatial resolution approximately 0.3 meters per pixel (equivalent to 1 foot), providing detailed visual information suitable for fine-grained scene classification tasks.

\subsubsection{AID}
The Aerial Image Dataset (AID) \cite{Xia_2017} includes 10,000 images from Google Earth, 600×600px, with varying resolutions (0.5–8 meters/px), spread across 30 classes with slight imbalance in distribution.

\subsubsection{NWPU}
Remote Sensing Image Scene Classification\cite{7891544}, created
by Northwestern Polytechnical University (NWPU-RESISC45) is an aerial imaging dataset that has 31,500 images, 256×256px, from global regions via Google Earth, covering 45 balanced categories with resolutions from 0.2 to 30 meters/px.

\subsection{Data preprocessing}

We normalize the training and validation data with mean $[0.485, 0.456, 0.406]$ and standard deviation $[0.229, 0.224, 0.225]$ for the rgb image channels.
Images are resized to 244x244 pixels for the vision transformers during the training pipeline, no matter their initial size.
\begin{table}[t]
\centering
\caption{Architecture details of the used ViT backbones. The table reports the patch size, number of transformer layers, hidden embedding size, and total parameter count (in millions) for each model variant. The parameter count for ViT-G14 is only for OReole given that there is no ViT-G14 for ImageNet, denoted by an asterisk.}
\label{tab_vits}

{
\begin{tabular}{@{}lcccc@{}}
\toprule
\textbf{Model} & \textbf{Patch Size} & \textbf{Layers} & \textbf{Hidden Size} & \textbf{Total Parameters [Millions]} \\
\midrule
ViT-B16    & 16 & 12 & 768  & 64.33 \\
ViT-H14    & 14 & 32 & 1280 & 472.67 \\
ViT-G14 & 14 & 32 & 1536 & 680.45* \\
\bottomrule
\end{tabular}
}
\end{table}
\begin{table}[t]
\centering
\caption{Constant hyperparameters used across all sweep experiments. These values remained fixed for every model, dataset, and training configuration to ensure consistency and comparability of results. The listed parameters include both general training settings and method-specific configurations for DLRT}
\label{tab:const_hyperparameters}
{
\begin{tabular}{@{}ll@{}}
\toprule
\textbf{Hyperparameter} & \textbf{Value} \\
\midrule
training batch size              & 16 \\
validation batch size              & 256 \\
training epochs                       & 10 \\
\midrule
initial rank (for truncated SVD)                      & 100 \\
maximum rank for DLRT augmentation                            & 200 \\
coefficient finetuning epochs (DLRT, LoRA)       & 2 \\
\bottomrule
\end{tabular}
}
\end{table}

\section{Model Details}
We refer to \cite{dosovitskiy2021imageworth16x16words} for the full description of the vision transformer architecture and display the architecture differentiators for the models used in this work in \Cref{tab_vits}.

\section{Training Details}
We display the hyperparameters, that remain constant across all runs in \Cref{tab:const_hyperparameters}. 
The lower part of the table denotes DLRT hyperparameters. We remark that in the last two epochs of training, we freeze the low-rank bases $U,V$ and only train $S$. This ensures that in the final steps of DLRT or LoRA, a sudden basis change does not affect validation performance. This trick is consistent with the loss, convergence and error bound analysis of DLRT \cite{zangrando2024geometryawaretrainingfactorizedlayers}. In practice, one does not expect much changes in the low-rank basis after a certain number of augmentation-truncation steps \cite{Coquelin_2024}.

The results of the hyperparameter sweeps can be found in \Cref{tab:combined_hyperparameters}. We perform a random sweep of at least 200 runs per model/data/method combination, where we vary the choice of weight-decay, learning rate, as well as local iterations and relative truncation tolerance $\vartheta=\tau\|W_r\|_F$.
We chose the same hyperparameters for LoRA and DLRT.

\begin{table}[t]
\caption{Hyperparameter sweep results by dataset, method, and model. 
$\gamma$ = weight decay, $\lambda$ = learning rate, $s_*$  = number of local iterations for the $S$ update on the tangent space. 
The threshold is defined as $\tau = \vartheta \cdot \|\,[\varsigma_{r_1}, \dots, \varsigma_{2r}]\,\|_2$.}

\label{tab:combined_hyperparameters}
\resizebox{\textwidth}{!}{
\centering
\setlength{\tabcolsep}{4pt}
\renewcommand{\arraystretch}{1.2}
\begin{tabular}{ll
                *{4}{c}
                *{4}{c}
                *{4}{c}
                                }
\toprule
& \multirow{2}{*}{\textbf{Model}} &
\multicolumn{4}{c}{\textbf{UCM}} &
\multicolumn{4}{c}{\textbf{AID}} &
\multicolumn{4}{c}{\textbf{NWPU}} \\
\cmidrule(lr){3-6} \cmidrule(lr){7-10} \cmidrule(lr){11-14}
& & $\gamma$ & $\lambda$ & $s_*$  & $\tau$ & $\gamma$ & $\lambda$ & $s_*$ & $\tau$  &$\gamma$ & $\lambda$ & $s_*$  & $\tau$   \\
\midrule
\multirow{4}{*}{\rotatebox[origin=c]{90}{ImageNet1k}}
& ViT-B16 Baseline & 0.00062 & 9.64e-5 &-- &--& 0.04055 & 8.66e-5 &-- &--& 0.04120 & 5.58e-5&-- &-- \\
& ViT-B16 Low-Rank       & 0.02461 & 1.33e-4 &181&0.0176& 0.03092 & 2.59e-4 &236 & 0.0789& 0.02692 & 9.13e-5 & 36 & 0.0761 \\
& ViT-H14 Baseline & 0.01437 & 2.49e-4 &-- &--&0.02637 & 3.49e-4&-- &-- & 0.03377 & 2.71e-4 & -- & -- \\
& ViT-H14 Low-Rank      & 0.00283 & 5.85e-4 &232 & 0.0497&0.04137 & 2.88e-4 &89&0.0081& 0.00554 & 4.79e-4 & 57  &0.0570 \\
\midrule
\multirow{6}{*}{\rotatebox[origin=c]{90}{OReole-MR}}
& ViT-B16 Baseline & 0.03720 & 1.09e-4 &-- &--& 0.00201 & 8.55e-5 &-- &--& 0.01662 & 6.69e-5 & -- & -- \\
& ViT-B16 Low-Rank & 0.03724 & 1.66e-4 &22&0.0379& 0.03156 & 1.81e-4 &64&0.0073& 0.02205 & 1.98e-4 & 226  & 0.0509 \\
& ViT-H14 Baseline & 0.02246 & 7.40e-5 &-- &--& 0.02222 & 7.37e-5 &-- &--& 0.00146 & 8.96e-5 & -- & -- \\
& ViT-H14 Low-Rank & 0.03318 & 2.01e-4 &119&0.0085& 0.01318 & 1.94e-4 &221&0.0393& 0.03547 & 3.16e-4 & 248 & 0.0229 \\
& ViT-G14 Baseline & 0.01121 & 6.97e-5 &-- &--& 0.01813 & 6.32e-5 &-- &--& 0.04006 & 8.38e-5 & -- & -- \\
& ViT-G14 Low-Rank & 0.04661 & 2.35e-4 &24&0.0235& 0.02901 & 9.03e-5 &160&0.0295& 0.04897 & 1.28e-4 & 113  & 0.0326 \\
\bottomrule
\end{tabular}
}
\end{table}

\begin{table}[t]
\caption{Correlation of tolerance, local iteration, learning rate, and weight decay with validation accuracy for the hyperparameter sweep. Negative correlations for learning rate across all models confirm that smaller values are preferred for stable training.}
\label{tab:hyperparam_correlation}
\centering
{

\setlength{\tabcolsep}{4pt}
\renewcommand{\arraystretch}{1.2}
\begin{tabular}{@{}llcccc@{}}
\toprule
 & \textbf{Model} & $\tau$ &$s_*$ & $\lambda$ & $\gamma$ \\
\midrule

\multirow{4}{*}{\rotatebox[origin=c]{90}{\shortstack{ImageNet21k}}}
  & ViT-B16 (Full Rank)          & - & - & -0.864 & -0.042 \\
  & ViT-H14 (Full Rank)          & - &  - & -0.818 & -0.052 \\
  & ViT-B16 (Low Rank)          & -0.123 & -0.024 & -0.932 & 0.034 \\
  & ViT-H14 (Low Rank)          & -0.023 & 0.03 & -0.895 & 0.025 \\
  \midrule
  \multirow{4}{*}{\rotatebox[origin=c]{90}{\shortstack{OReole-MR}}}
  & ViT-B16 (Full Rank)          & - & - & -0.83 & 0.038 \\
  & ViT-H14 (Full Rank)          & - &  - & -0.885 & 0.094 \\
  & ViT-B16 (WLR)                &  0.081 & 0.026 & -0.807 & 0.042 \\
  & ViT-H14 (WLR)                & -0.012 & 0.009 & -0.876 & -0.096 \\

\bottomrule
\end{tabular}
}
\end{table}

\section{Hardware}
All trainings and sweeps were run on single workstation GPUs, using NVIDIA RTX A6000 40GB and  NVIDIA RTX 4090 24GB GPUS.

\end{document}